\pdfoutput=1
\documentclass[11pt]{article}

\usepackage{CJK}
\usepackage[preprint]{acl}

\usepackage[varqu,varl]{inconsolata}

\usepackage{times}
\usepackage{makecell}
\usepackage{arydshln}
\usepackage{latexsym}
\usepackage[utf8]{inputenc}
\usepackage{booktabs} 
\usepackage[T1]{fontenc}

\usepackage{hyperref}  
\usepackage{amsfonts} 
\usepackage{microtype}
\usepackage{xcolor}  
\usepackage{inconsolata}

\usepackage{graphicx}
\usepackage{multirow}
\usepackage{array}
\usepackage{xspace}

\usepackage{amssymb}
\title{COIG-CQIA: Quality is All You Need for Chinese Instruction Fine-tuning}

\author{
    \textbf{Yuelin Bai}\textsuperscript{1}\thanks{\ \ Equal contribution.}\quad
    \textbf{Xinrun Du}\textsuperscript{2}\footnotemark[1]\quad
    \textbf{Yiming Liang}\textsuperscript{3}\footnotemark[1]\quad
    \textbf{Yonggang Jin}\textsuperscript{2}\footnotemark[1]
    \textbf{Junting Zhou}\textsuperscript{2,4}\footnotemark[1]\quad\\
    \textbf{Ziqiang Liu}\textsuperscript{1}\quad
    \textbf{Feiteng Fang}\textsuperscript{5}\quad
    \textbf{Mingshan Chang}\textsuperscript{1}\quad
    \textbf{Tianyu Zheng}\textsuperscript{2}\quad
    \textbf{Xincheng Zhang}\textsuperscript{5}\quad\\
    \textbf{Nuo Ma}\textsuperscript{6}
    \textbf{Zekun Wang}\textsuperscript{2}\quad
    \textbf{Ruibin Yuan}\textsuperscript{2,7}\quad
    \textbf{Haihong Wu}\textsuperscript{5}\quad
    \textbf{Hongquan Lin}\textsuperscript{5}\quad\\
    \textbf{Wenhao Huang}\textsuperscript{6}
    \textbf{Jiajun Zhang}\textsuperscript{3}\quad
    \textbf{Chenghua Lin}\textsuperscript{2,10}\quad
    \textbf{Jie Fu}\textsuperscript{7}\quad
    \textbf{Min Yang}\textsuperscript{1}\\
    \textbf{Shiwen Ni}\textsuperscript{1}\thanks{\ \ Corresponding authors.}\quad
    \textbf{Ge Zhang}\textsuperscript{8,9}\footnotemark[2]\\    
    \textsuperscript{1}Shenzhen Institute of Advanced Technology, CAS\quad
    \textsuperscript{2}M-A-P \quad 
    \textsuperscript{3}Institute of Automation, CAS\\
    \textsuperscript{4}Peking University\quad
    \textsuperscript{5}University of Science and Technology of China\quad
    \textsuperscript{6}01.ai\quad
    \textsuperscript{7}HKUST\\
    \textsuperscript{8}University of Waterloo\quad
    \textsuperscript{9}Vector Institute\quad
    \textsuperscript{10}University of Manchester\quad
    \vspace{10mm}
\small \\
}

\newcommand{\data}{\texttt{COIG-CQIA}\xspace}

\definecolor{navyblue}{rgb}{0.0, 0.0, 0.5}
\definecolor{oceanboatblue}{rgb}{0.0, 0.47, 0.75}
\definecolor{persianblue}{rgb}{0.11, 0.22, 0.73}
\definecolor{persianred}{rgb}{0.8, 0.2, 0.2}

\begin{document}
\begin{CJK}{UTF8}{gbsn}

\maketitle

\begin{abstract}
Remarkable progress on English instruction tuning has facilitated the efficacy and reliability of large language models (LLMs).
% Remarkable progress on large language models (LLMs), particularly in English, has facilitated impressive capabilities in following human instructions.
% impressive accuracy and fluency in understanding and following human instructions.
However, there remains a noticeable gap in instruction tuning for Chinese, where the complex linguistic features pose significant challenges. Existing datasets, 
% generally derived from English-centric LLMs or NLP datasets,
generally distilled from English-centric LLMs, are not well-aligned with Chinese users' interaction patterns.
To bridge this gap, we introduce \data, a new Chinese instruction tuning dataset derived from various real-world resources and undergoing rigorous human verification.
% composed of human-written instruction-output pairs derived from wide-ranging real-world data resources.} 
% Our aim is to build a high-quality dataset facilitating LLMs' behaviors better aligned with human interactions. 
% We establish a corpus of high-quality human-written instruction-output pairs from various Chinese Internet sources, covering Social media \& Forums, world knowledge, examinations, and existing NLP datasets. 
% This corpus was rigorously filtered and processed to form \data.
We conduct extensive experiments on \data, and compare them with strong baseline models and datasets.
The experimental results show that models trained on \data achieve highly competitive performance in diverse benchmarks.
% Models trained on \data achieve highly competitive results in human assessment, instruction-follow, knowledge, and safety benchmarks.
% 改为：Models trained on our dataset show proficiency in following Chinese instructions.
Additionally, our findings offer several insights for designing effective Chinese instruction-tuning datasets and data-mixing strategies.
Our dataset are available at \url{https://huggingface.co/datasets/m-a-p/COIG-CQIA}.

\end{abstract}

\section{Introduction}
\label{intro}
% 缺少中文数据集，更缺少对各来源对downstream task的影响的分析，我们在构建数据过程中，investigate这一点

Large Language Models (LLMs)
% , such as GPT-3  \citep{brown2020language}, LLaMA  \citep{touvron2023llama}, and PaLM  \citep{chowdhery2023palm}, 
have shown remarkable capabilities as general-purpose assistants.
The cornerstone of this advancement is instruction tuning  \citep{zhang2023instruction}, which significantly enhances the efficacy and safety of models in following human instructions.
The core idea is to train models with instruction-output pair data, thus aligning the model's training objective with human intent.
% significantly enhances the capabilities and controllability of LLMs through training on datasets composed of instruction-output pairs.
% This technique effectively aligns the models' training objectives with human intentions, thus enhancing the efficacy and safety in following human instructions.
% thereby ensuring that the models can interpret and execute human instructions both effectively and safely.
% Therefore, the availability of high-quality instruction tuning datasets is crucial for LLMs to operate as efficient and dependable assistants.
This highlights the key role of high-quality instruction tuning datasets in enabling LLMs to function as efficient and reliable assistants.
Despite the remarkable progress made in English instruction tuning datasets, datasets for Chinese instruction tuning still remain in the nascent stages. 
% So far, substantial progress has been made on English instruction tuning. However, Chinese instruction tuning is still in the nascent stage.
% Existing datasets for Chinese instruction tuning generally suffer from limited scale or suboptimal quality. 
Existing datasets can be roughly categorized into three types:
(1) Datasets derived from English instruction datasets  \citep{peng2023instruction} or NLP datasets \citep{coig-pc, Firefly}, 
(2) Datasets synthesized by LLMs \citep{guo2023close,ji2023exploring,sun2023moss}, and 
(3) Hybrid dataset constructed using different methods \citep{zhang2023chinese}.
% These datasets generally suffer from limited scale or suboptimal quality. 
To improve dataset quality, COIG \citep{zhang2023chinese} leveraged multiple methods to construct a human-verified instruction corpus.
% , aiming to address the shortcomings of its predecessors.
% However, prior Chinese instruction-tuning datasets still face several challenges: 
However, two major challenges still exist in prior Chinese instruction tuning datasets. First, they suffer from insufficient alignment with real-world Chinese users due to the lack of authentic linguistic data. Second, they are still riddled with quality issues due to the high cost of comprehensive human verification.
% still suffer from limited scale or suboptimal quality. 
% they often read as unnatural to native Chinese speakers, lack authentic linguistic data, 
% are riddled with quality issues, and remain limited in scale. 
Moreover, it is still under-explored how different data sources impact the downstream Chinese tasks, exacerbating the challenges in constructing Chinese datasets.
% remaining a gap in understanding the influence of data sources on the model performance. 
% This knowledge gap consequently makes the construction of Chinese datasets even more challenging.

\begin{figure*}[ht]
    \centering
    \resizebox{0.95\textwidth}{!}{\includegraphics{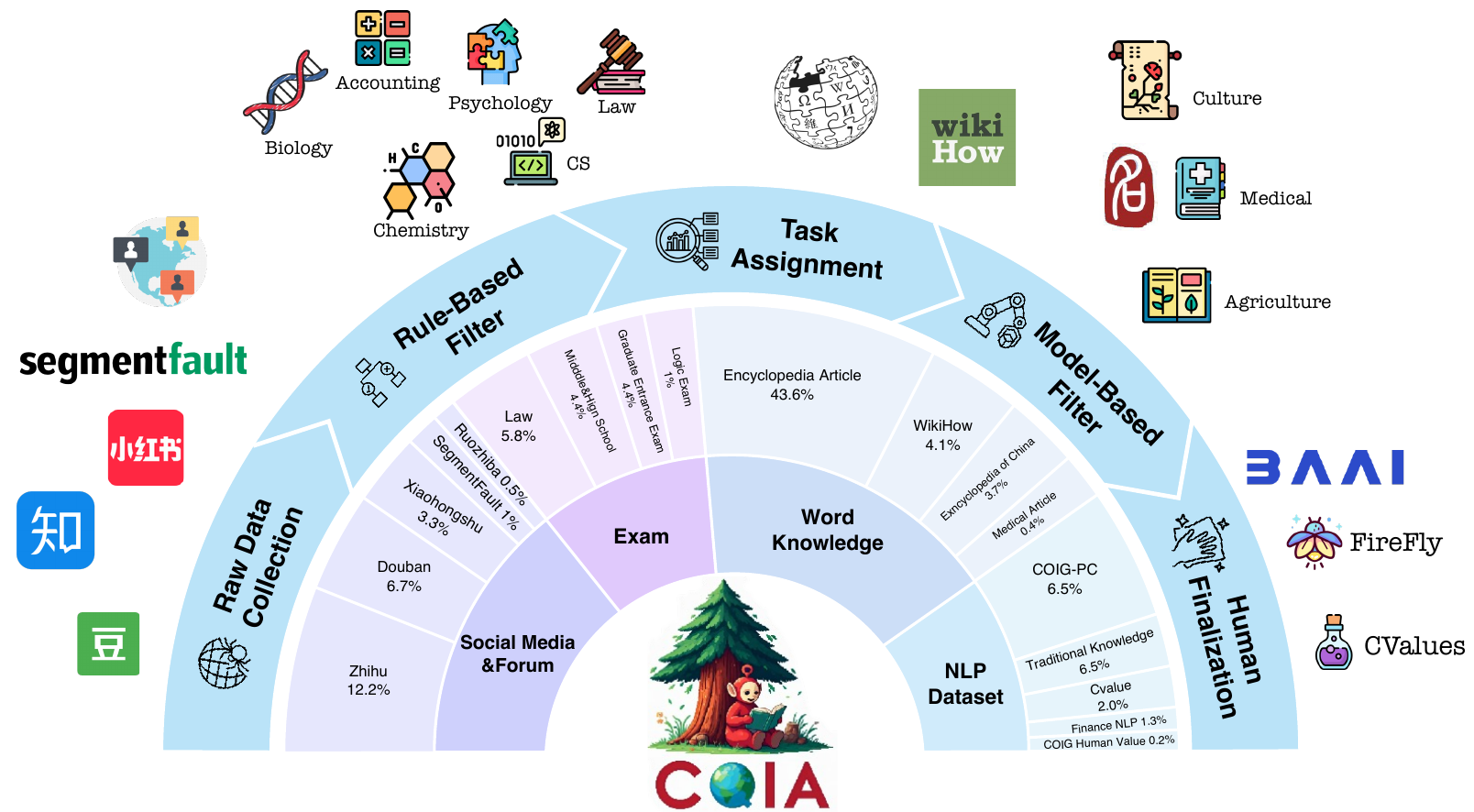}} 
    \caption{Overview of \data and Statistics of each data source.} 
    \label{fig:main_fig} 
\end{figure*}

To address these challenges, we introduce \data~(\textbf{C}hinese \textbf{O}pen \textbf{I}nstruction \textbf{G}eneralist - \textbf{Q}uality \textbf{I}s \textbf{A}ll You Need), a new Chinese instruction tuning dataset, 
distinguished by its incorporation of diverse real-world data resources and rigorous human verification processes.
% , aiming to enhance the quality and authenticity of Chinese language instruction datasets.
% derived from rich real-world data resources and meticulous human verification.
% designed to provide the NLP community with instruction-tuning data that not only meets high quality but also aligns well with real human interactions.
Inspired by LIMA \citep{zhou2023lima}, \data focuses on curating a dataset from diverse Chinese internet sources, covering social media and forums, comprehensive encyclopedias, challenging examinations, and existing linguistic corpus. 
These data undergo a thorough cleaning, restructuring, and careful human verification to secure the quality and diversity. 
Our aim is to enhance the proficiency of LLMs in following Chinese instructions and executing downstream tasks.
% Through this process, we aim to enhance the proficiency of LLMs in following Chinese instructions and improve their overall performance on downstream tasks.
We conduct extensive experiments to explore how different data sources impact various downstream tasks and explored the benefits of different data mixing strategies. Additionally, we integrate \data with English data to investigate the performance of trained models in multilingual scenarios.
% instruction-tuning tasks.
Further experiments show that \data achieves highly competitive results compared to other Chinese datasets and strong baseline models. Our main contributions are as follows:
% Further experiments show that \data is highly competitive in comparison with other baseline Chinese datasets and strong baseline models.

% \begin{itemize}
%     \item We propose a high-quality Chinese instruction-tuning dataset \data, specifically designed to align with human interaction, achieved through rigorous filtering procedures.
%     \item %We investigate the influence of diverse data sources, such as social media, encyclopedias, and conventional NLP tasks, on model performance. Our analysis provides insights crucial for selecting instruction data from the Chinese internet.
%     We explore the influence of various data sources, including social media, encyclopedias, and traditional NLP tasks, on model performance. Our analysis offers essential insights for selecting training data from the Chinese internet.
%     \item Various benchmark tests and human evaluations confirm that models fine-tuned on our \data exhibit superior performance, thus establishing \data as a valuable resource for the Chinese NLP community.
% \end{itemize}

\begin{itemize}
    \item \textbf{Resource.} We introduce \data, a high-quality Chinese instruction-tuning dataset built from diverse, real-world sources and verified by humans to ensure quality.
    \item \textbf{Performance.} We demonstrate the effectiveness of \data through extensive experiments, showing its competitiveness against other Chinese datasets and baseline models.
    \item \textbf{Insights.} We systematically investigate the impact of different data sources and mixing strategies on downstream task performance, providing insights into data source influence and training strategy.
\end{itemize}

\section{COIG-CQIA CURATION}

To ensure data quality and diversity, we curated data from 18 high-quality Chinese Internet sources. We also integrated existing Chinese NLP datasets and examinations to broaden task diversity. Specifically, we categorized all data sources into four types: Social Media \& forums, World Knowledge, NLP tasks, and Exam. The statistical information of the data are detailed in the table \ref{tab:CQIA_source}.

\newcommand{\coloredblock}[1]{\textcolor{#1}{\blacksquare}}
\definecolor{raw_data_acquisition}{rgb}{0.0, 0.5, 1.0} 
\definecolor{rule_based}{rgb}{0.0, 0.8, 0.0}       
% \definecolor{task_classification}{rgb}{1.0, 0.5, 0.0} 
\definecolor{template_design}{rgb}{1.0, 0.5, 0.0}  
\definecolor{model_selection}{rgb}{1.0, 0.0, 0.5}     
\definecolor{manual_selection}{rgb}{0.5, 0.0, 0.5}     

\begin{table*}[ht]
\centering
\resizebox{\textwidth}{!}{
\begin{tabular}{ccccl}
    \toprule
    \textbf{Source} & \textbf{Type} & \textbf{Description} & \textbf{Quantity} & \textbf{Data Processing}\\ 
    \midrule
    Zhihu $(\ref{appdix:Zhihu})$ & Forum & Comprehensive Q\&A platform & 8837 &  $\coloredblock{raw_data_acquisition}$ $\coloredblock{rule_based}$ $\coloredblock{model_selection}$ $\coloredblock{manual_selection}$ \\ 
    Segment Fault (\ref{appdix:SegFault}) & Forum & Tech community for IT developers & 458 & $\coloredblock{raw_data_acquisition}$ $\coloredblock{rule_based}$ $\coloredblock{manual_selection}$ \\ 
    Douban (\ref{appdix:Douban}) & Social Media & User-driven platform focused on literature and the arts & 3132 & $\coloredblock{raw_data_acquisition}$ $\coloredblock{rule_based}$ 
    $\coloredblock{template_design}$
    $\coloredblock{manual_selection}$ \\ 
    Xiaohongshu (\ref{appdix:Xiaohongshu}) & Social Media & Life experiences sharing platform & 1508 & $\coloredblock{raw_data_acquisition}$ $\coloredblock{rule_based}$ 
    $\coloredblock{template_design}$
    $\coloredblock{manual_selection}$ \\ 
    Ruozhiba (\ref{appdix:Ruozhiba})  & Forum & Tieba\footnote{China’s largest interest-based online community platform} subcommunity interested in logical traps. & 240 & $\coloredblock{raw_data_acquisition}$ $\coloredblock{rule_based}$ 
    $\coloredblock{model_selection}$ $\coloredblock{manual_selection}$ \\ 
    \midrule
    Encyclopedia Article & World Knowledge & Comprehensive encyclopedic knowledge from various website & 20020 & $\coloredblock{raw_data_acquisition}$ $\coloredblock{rule_based}$ 
    $\coloredblock{template_design}$
    $\coloredblock{manual_selection}$ \\ 
    Encyclopedia of China(\ref{appdix:GeneralEncyclopedia}) & World Knowledge & Comprehensive Chinese encyclopedia & 1706 & $\coloredblock{raw_data_acquisition}$ $\coloredblock{rule_based}$ 
    $\coloredblock{template_design}$
    $\coloredblock{manual_selection}$ \\ 
    WikiHow (\ref{appdix:wikihow})& World Knowledge & Step-by-step guides and how-tos & 1876 & $\coloredblock{raw_data_acquisition}$ $\coloredblock{rule_based}$ 
    $\coloredblock{manual_selection}$ \\ 
    Medical Ariticle & World Knowledge & Health-related knowledge & 186  & $\coloredblock{raw_data_acquisition}$ $\coloredblock{rule_based}$ 
    $\coloredblock{manual_selection}$ \\ 
    \midrule
    Middle\&High School Exam(\ref{appdix:mid_school_exam}) & Exam & Standardized exam for middle\&high school students & 2000 & $\coloredblock{raw_data_acquisition}$ $\coloredblock{rule_based}$ 
    $\coloredblock{manual_selection}$ \\ 
    Graduate Entrance Exam(\ref{appdix:Graduate Entrance Exam}) & Exam & National graduate entrance exam & 475 & $\coloredblock{raw_data_acquisition}$ $\coloredblock{rule_based}$ 
    $\coloredblock{manual_selection}$ \\ 
    Logical Exam(\ref{appdix:Logical exam}) & Exam & Logistic reasoning exam questions & 422 & $\coloredblock{raw_data_acquisition}$ $\coloredblock{rule_based}$ 
    $\coloredblock{manual_selection}$ \\ 
    Law Exam & Exam & Law graduate entrance exam questions & 2645 & $\coloredblock{raw_data_acquisition}$ $\coloredblock{rule_based}$ 
    $\coloredblock{manual_selection}$ \\ 
    \midrule
    COIG PC(\ref{appdix:COIG_PC}) & NLP Dataset & A massive dataset for instruction fine-tuning & 3000 & $\coloredblock{rule_based}$ 
    $\coloredblock{manual_selection}$ \\ 
    COIG-Human-Value(\ref{appdix:COIG_Human_value}) & NLP Dataset & Value-related tasks from COIG & 101 & $\coloredblock{rule_based}$ 
    $\coloredblock{manual_selection}$ \\ 
    CValues(\ref{appdix:CValue}) & NLP Dataset & Detoxifying answers written by experts for harmful questions & 906 & 
    $\coloredblock{manual_selection}$ \\ 
    Chinese Traditional (\ref{appdix:chinese culture})& NLP Dataset & Chinese traditional culture tasks from various datasets & 1111 & $\coloredblock{raw_data_acquisition}$ $\coloredblock{rule_based}$ 
    $\coloredblock{manual_selection}$ \\ 
    Finance NLP Task & NLP Dataset & NLP tasks in the financial domain & 600 & $\coloredblock{rule_based}$ 
    $\coloredblock{manual_selection}$ \\
    \midrule
    \textbf{Total} & & & \textbf{45173} \\ 
    \bottomrule
\end{tabular}
}
\caption{Overview of different data sources. We list each data source's source type, description, quantity, and data processing stages. The colored squares represent different stages we applied to the data processing:
$\coloredblock{raw_data_acquisition}$ Raw Data Collection; 
$\coloredblock{rule_based}$ Rule-based filter; 
% $\coloredblock{task_classification}$ Task Assignment; 
$\coloredblock{template_design}$ Template Curation; 
$\coloredblock{model_selection}$ Model-Based Filter; 
$\coloredblock{manual_selection}$ Human Finalization.
}
\label{tab:CQIA_source}
\end{table*}

% \begin{figure}[h]
%     \centering
%     \resizebox{0.95\columnwidth}{!}{\includegraphics{fig/instruction_output_length_distributions.pdf}} 
%     \caption{Length distribution of instruction and responses. Note that the instruction is the concatenation of instruction and input of each data instance.}\byl{move to appendix}
%     \label{fig:sequence_length} 
% \end{figure}

\subsection{Raw Data Collection}
In this initial stage, we aggregated data from diverse sources from the Internet, including social media platforms, encyclopedias, and specialized websites. For instance, we used a web crawler to collect posts from Zhihu, SegmentFault, etc., as well as entries from encyclopedic sources such as the Encyclopedia of China. For the crawled HTML content, we carefully converted them into question-answer pairs or documents. We preserved as much non-text metadata as possible, such as likes, comments, authors, multimedia elements, etc., to facilitate rule-based filtering based on this metadata.
Meanwhile, we collected publicly available official exam papers from previous years and used the Mathpix tool \footnote{https://mathpix.com/} to extract the questions and detailed answers from the documents.
In Table \ref{tab:CQIA_source}, we mark all the sources that are processed into plain text from crawled or non-text corpora using $\coloredblock{raw_data_acquisition}$.

\subsection{Rule-Based Filter}

The rule-based filter is is implemented subsequent to the extraction of plain text data. Its primary purpose is to perform preliminary data cleansing, eliminating content that contains harmful or inappropriate information,  multimedia elements, or advertisements. Additionally, it removes data that fails to meet specified length criteria or lacks wide human acceptance (e.g., posts with very few likes in forums). This stage is essential while highly efficient, capable of reducing the dataset from millions of entries to hundreds of thousands. Rule-based filtering was applied to nearly all data sources, as marked by green squares $\coloredblock{rule_based}$ in the table \ref{tab:CQIA_source}.  

\subsection{Task Assignment}

This stage is used to convert raw posts and articles into instruction-response formats for instruction fine-tuning. Overall, we designed a variety of instruction or response templates based on the characteristics of the content from different data sources. For example, for encyclopedic entries, the tasks focus mainly on concept explanation, while for metadata-rich sources like Douban, we designed tasks around reviews writing, recommendation, etc. All the details of the data construction process are described in the appendix \ref{app:data curation}. Not all data sources require manually designed templates, and those are marked with orange squares $\coloredblock{template_design}$ in Table 1.

\subsection{Model-Based Filter}

While data from social media and forums closely reflects real human interactions and offers great diversity, it's challenging to ensure that all of this data is harmless and accurate. Model-based filtering (marked by pink squares $\coloredblock{model_selection}$) can help eliminate low-quality data that's difficult to remove through rule-based or metadata filtering. This typically includes irrelevant instruction-response pairs, soft advertisements, and potentially harmful content. We used GPT-4 to filter the data sources, as it's widely used in LLMs-as-judge and demonstrates a high correlation with human judgment in assessing data quality. We detail this process for specific sources in the appendix \ref{app:data curation}.

\subsection{Human Finalization}

To ultimately ensure data quality, we invite human reviewers to re-examine all the data and finalize each data source. Consistent with our model-based filtering criteria, we asked human judges to evaluate the data's usefulness, professionalism, logical coherence, level of detail, objectivity, and harmlessness. Given the flexibility and variability of Chinese language usage, human review allows for filtering out cases that rule-based and model-based approaches struggle to identify. This process also involves making appropriate modifications to instruction-response pairs to ensure accuracy and alignment between responses and the instructions' intent. We detail this process in the appendix \ref{app:data curation}.

\section{Data Analysis}
\vspace{-0.1cm}
\subsection{Statistics} We collected a total of 45,173 instances from 18 sources within the Chinese Internet and Community, covering domains ranging from general knowledge and STEM to humanities. 
Table \ref{tab:CQIA_source} describe the data statistics for all sources.
We demonstrated the distribution in the length of the instructions and responses in Figure \ref{fig:sequence_length}.
\vspace{-0.1cm}
\subsection{Semantic Distribution} 
To visualize and analyze the semantic diversity of our dataset, we employed U-MAP to create a distribution map of all the instructions. Figure \ref{fig:u-map} in Appendix \ref{appdix:comparison} illustrates the semantic distribution of \data compared to other datasets. The U-MAP visualization reveals that \data exhibits the most widespread and diverse distribution among all compared datasets. 
\vspace{-0.1cm}
\subsection{Quality}
We sample a total of 100 data instances from all the sources within the dataset, and then manually evaluate their quality based on four criteria: (1) Is the output correct and an acceptable answer? (2) Does the output meet the instructional requirements and provide a comprehensive and appropriate response to the question? (3) Is the answer complete and sufficiently detailed? (4) Is the answer harmless, avoiding misleading information or the spread of harmful content?

Our human evaluation results in table \ref{tab:acc_rate} shows that the data quality has met a very high standard, with human acceptance rates consistently above 95\% across four criteria. Regarding the third criteria, we conducted a case study which reveal that human rejections were primarily due to responses not being excessively detailed\footnote{Such as how AI systems like GPT4 would include redundant information in their responses}. Given that our responses are primarily collected from real human interactions on the web, we believe it's acceptable and even natural.

% \subsection{Diversity} To analyze the diversity of the COIG-CQIA dataset, we follow prior work\cite{wang2023selfinstruct, lou2023muffin} by employing the Hanlp tool\cite{he2021stem} to parse the instructions and then extract the verb closest to the root along with its top direct noun object.\footnote{We only visualize when a certain verb-noun pair has more than 30 instances, and many instructions do not contain a verb-noun structure.} We then plot the top 20 most common root verbs and their corresponding direct noun objects in Figure\ref{fig:root_verb}. From this figure we can observe that CQIA features a diverse range of instructions and intentions.

\vspace{-0.2cm}
\section{EXPERIMENTAL SETUP}

In this section, we describe how we use \data to fine-tune models and elaborate our evaluation methods.

\paragraph{Evaluation Benchmarks} To assess the model's capabilities across various Chinese tasks, we utilize Belle-Eval \citep{ji2023exploring} as our open-ended test set. It encompasses 12 different instruction types spanning various domains, making it ideal for evaluating the impact of different data sources on various tasks. We also employed C-Eval \citep{huang2024c}, CMMLU \citep{li2023cmmlu}, and SafetyBench \citep{zhang2023safetybench}, which are widely used benchmarks for assessing models' knowledge, reasoning, and safety levels in Chinese contexts.
To further explore \data's extensibility in non-Chinese scenarios, we evaluated the model on widely-used datasets including BBH \citep{suzgun2022challenging}, GSM8K  \citep{cobbe2021gsm8k}, HumanEval \citep{chen2021evaluating}, and TydiQA \citep{clark2020tydi}.

\paragraph{Baselines} 
To comprehensively evaluate the instruction-following capacity of the models fine-tuned on \data, we compared it with several well-known Chinese instruction-tuning datasets.
These include COIG \citep{zhang2023chinese}, Firefly \citep{Firefly}, Alpaca-ZH \citep{chinese-llama-alpaca}, COIG-PC \citep{coig-pc}, and OL-CC \citep{olcc}, which were constructed using various methods. We compared subsets of equal size to \data from these datasets. Additionally, we chose subsets totaling the same size as \data -Sub from WizardCoder \citep{luo2023wizardcoder} and MAmmoTH \citep{yue2023mammoth} for data mixing experiments.\footnote{See section \ref{sec:os mix} for details.}

\begin{table*}[ht]
\centering
\resizebox{0.96\textwidth}{!}{%
\begin{tabular}{clccccccccccc}
    \toprule
    & \textbf{Dataset}  & \textbf{Open-QA} & \textbf{Brain.} & \textbf{CLS.} & \textbf{Gen.} & \textbf{Sum.} & \textbf{Rewrite} & \textbf{Closed-QA} & \textbf{Extract} & \textbf{Math} & \textbf{Code} & \textbf{Average} \\
    \midrule
    \multicolumn{13}{c}{\textit{\textbf{Vanilla Models}}} \\
    \midrule
    \multirow{2}{*}{\rotatebox[origin=c]{90}{\textbf{}}} & Vanilla Qwen-2-7B & 65.5 & 60.0 & 46.0 & 54.3 & 40.7 & 53.5 & 58.7 & 44.5 & 46.2 & 67.1 & 53.7 \\
    & Vanilla LLaMA-2-13B & 1.4 & 3.8 & 5.0 & 1.0 & 6.7 & 17.5 & 12.2 & 13.6 & 0.0 & 17.1 & 6.9 \\
    \midrule
    \multicolumn{13}{c}{\textit{\textbf{Qwen2-7B trained on different \data\ data source}}} \\
    \midrule
    \multirow{15}{*}{\rotatebox[origin=c]{90}{\textbf{Qwen-2-7B}}} & Zhihu & 65.2 & 89.6 & 42.0 & 91.9 & 42.7 & 56.5 & 36.1 & 37.3 & 77.6 & 80.0 & 63.7 \\
    & Douban & 53.8 & 67.3 & 15.0 & 68.1 & 13.3 & 34.0 & 37.8 & 27.3 & 81.0 & 43.6 & 47.0 \\
    & Xhs & 49.3 & 60.0 & 12.5 & 42.9 & 13.3 & 12.0 & 31.7 & 16.4 & 71.4 & 27.1 & 36.9 \\
    & SegmentFault & 53.8 & 68.5 & 41.5 & 69.0 & 33.3 & 74.5 & 48.7 & 42.7 & 76.2 & 65.7 & 58.6 \\
    & Ruozhiba & \textbf{77.6} & \textbf{95.8} & \textbf{64.5} & \textbf{96.7} & \textbf{76.7} & \textbf{91.5} & \textbf{82.6} & \textbf{72.3} & \textbf{90.5} & \textbf{87.1} & \textbf{83.5} \\
    & Exam & 51.4 & 83.8 & 54.2 & 75.2 & 30.7 & 73.0 & 72.2 & 57.3 & 49.5 & 71.4 & 62.9 \\
    & Logi QA & 52.1 & 69.2 & 50.5 & 78.6 & 25.3 & 70.0 & 53.7 & 50.0 & 75.7 & 65.7 & 60.2 \\
    & WikiHow & 48.3 & 28.5 & 1.0 & 41.9 & 20.7 & 5.0 & 20.9 & 12.7 & 62.4 & 47.9 & 30.2 \\
    & COIG PC & 53.1 & 95.4 & 53.0 & 85.2 & 47.3 & 56.5 & 50.4 & 60.0 & 61.9 & 42.9 & 62.1 \\
    & Chinese Tra & 41.7 & 73.1 & 41.0 & 79.5 & 28.7 & 69.5 & 55.2 & 41.8 & 80.0 & 58.6 & 58.2 \\
    % & Finance & 52.4 & 75.0 & 45.2 & 91.9 & 25.3 & 63.0 & 49.3 & 43.6 & 45.2 & 42.9 & 55.5 \\
    & Human Value & \underline{65.5} & 90.0 & \underline{60.5} & 86.7 & 58.0 & 85.0 & 64.8 & 50.9 & 78.6 & 72.9 & \underline{72.8} \\
    \cmidrule{2-13}
    & \data -Fullset & 63.8 & 88.3 & 55.0 & \underline{92.9} & 51.0 & 59.0 & \underline{67.8} & \underline{64.5} & 66.7 & 65.7 & 68.7 \\
    & \data -Subset & 59.7 & 86.2 & 54.0 & 91.9 & \underline{54.3} & 58.5 & 68.3 & 70.9 & \underline{83.3} & \underline{71.4} & 70.3 \\
    \midrule
    \multicolumn{13}{c}{\textit{\textbf{LLaMA-2-13B trained on different \data\ data source}}} \\
    \midrule
    \multirow{14}{*}{\rotatebox[origin=c]{90}{\textbf{LLaMA-2-13B}}} & Zhihu & 23.1 & 48.5 & 17.0 & 47.1 & 25.3 & 24.0 & 26.1 & 20.9 & 0.5 & 25.0 & 26.5 \\
    & Douban & 19.0 & 27.7 & 9.0 & 26.7 & 13.3 & 25.0 & 45.7 & 20.0 & 11.9 & 15.7 & 22.2 \\
    & Xhs & 15.9 & 28.5 & 0.0 & 23.8 & 6.7 & 25.0 & 25.2 & 20.9 & 1.0 & 10.0 & 16.3 \\
    & SegmentFault & 23.8 & 23.1 & 6.0 & 31.4 & 23.3 & 38.0 & 30.9 & 20.0 & 9.5 & 38.6 & 24.3 \\
    & Ruozhiba & 37.6 & 55.8 & \textbf{44.5} & 51.0 & \underline{39.3} & 38.5 & \textbf{55.2} & 34.1 & \underline{17.6} & \textbf{47.9} & \underline{42.7} \\
    % \cmidrule{2-13}
    & Exam & 30.7 & 60.0 & \underline{36.0} & 56.2 & 26.7 & 33.0 & 40.9 & 37.3 & 13.8 & 39.3 & 38.0 \\
    & Logi QA & 20.7 & 23.1 & 25.0 & 36.7 & 23.3 & 44.0 & 50.9 & \textbf{46.4} & 15.7 & 20.0 & 29.9 \\
    & Wiki & 25.5 & 52.3 & 15.0 & 50.0 & 10.0 & 17.5 & 31.7 & 43.6 & 4.8 & 40.7 & 29.1 \\
    & WikiHow & 26.6 & 24.2 & 5.0 & 34.3 & 10.0 & 15.0 & 21.7 & 9.1 & 2.4 & 28.6 & 18.6 \\
    & COIG PC & 22.8 & 28.8 & 22.5 & 22.4 & 23.3 & 32.5 & 40.4 & 23.6 & 7.1 & 12.1 & 24.2 \\
    & Chinese Traditional & 17.2 & 25.8 & 16.0 & 51.4 & 32.0 & 45.0 & 45.7 & 30.0 & 14.3 & 7.9 & 28.7 \\
    & Human Value & 33.4 & 61.9 & 35.0 & 64.3 & 25.3 & \textbf{49.0} & 40.0 & 46.4 & 4.3 & 33.6 & 39.9 \\
    \cmidrule{2-13}
    & \data -Fullset & \textbf{46.2} & \textbf{68.1} & 24.0 & \underline{65.2} & 25.3 & 36.5 & 43.5 & 39.1 & \textbf{18.6} & 35.0 & 41.9 \\
    & \data -Subset & 39.7 & 64.2 & 24.5 & \textbf{68.1} & \textbf{40.3} & \textbf{45.5} & 50.0 & \underline{45.5} & 8.1 & \underline{44.3} & \textbf{43.5} \\
    \bottomrule
\end{tabular}
}
\caption{\label{tab:belle}The performance of Qwen-2-7B and LLaMA-2-13B trained on various datasets evaluated on BELLE-EVAL using GPT-4o. Brain. refers to Brainstorm; CLS. refers to Classification; Gen. refers to generation; Sum. refers to summerization.}
\end{table*}

\paragraph{Implementation Details} 
We fine-tuned various models of different architectures and sizes using \data. This included Chinese-centric multilingual models from the Yi series (6B and 34B) \citep{young2024yi} and the Qwen2 series (7B and 72B) \citep{yang2024qwen2}. We merged the 18 data sources of the \data dataset into 12 sources, and manually selected a more balanced subset from these sources, which we refer to as \data -Sub. Detailed statistics can be found in the Appendix \ref{app:cqia-sub}. To explore the data's performance on non-Chinese-centric models, we also selected the LLaMA2 series (7B, 13B, and 70B) \citep{touvron2023llama} as base models.
We set the learning rate to 2e-5, with a batch size of 128 and a maximum sequence length of 4096. The training was conducted for 5 epochs using a cosine scheduler with 5\% warmup. For models under 20B parameters, we employ DeepSpeed \citep{rasley2020deepspeed} ZeRO stage 2 optimization, while for models larger than 20B parameters, we use ZeRO stage 3.

% \begin{table}[h]
% \centering
%     \resizebox{0.42\columnwidth}{!}{
%     \begin{tabular}{lc}
%     \toprule
%     Model          & SafetyBench \\ 
%     \midrule
%     COIG PC      & 81.2 \\
%     Chinese Tradiational  & 76.6 \\
%     Douban       & 76.2 \\
%     Exam   & 77.6 \\
%     Finance   & 75.1 \\
%     Logi QA & 79.1 \\
%     Ruozhiba & 81.3 \\
%     Segmentfault & 78.0 \\
%     Wiki & 75.8 \\
%     Wikihow & 76.4 \\
%     Xhs & 76.0 \\
%     Zhihu & 75.8 \\
%     Human Value & 79.1 \\
%     \midrule
%     CQIA-Sub-6B & \textbf{81.7} \\
%     \midrule
%     GPT-4-0613 & 89.2 \\
%     GPT-3.5-turbo-0613 & 80.4 \\
%     \bottomrule
% \end{tabular}
% }
% \vspace{1em}
% \caption{\label{tab:safetybench}SafetyBench scores of Yi-6B trained on various data sources.}
% \end{table}

% \begin{table}[h]
% \centering
% \resizebox{0.7\columnwidth}{!}{
% \begin{tabular}{lcc}
% \toprule
% Model          & Ceval (val 5-shot) & CMMLU (test 5-shot) \\ 
% \midrule
% Qwen-1.8b      & 51.34              & 47.26                   \\
% Yi-6B          & 73.40              & 74.85               \\
% Qwen-14b       & 68.20              & 67.96                   \\
% InternLM2-20b   & 71.25              & 67.48                   \\
% Yi-34b         & 77.04              & 78.18                   \\
% Qwen-72b       & 78.68              & 76.79                   \\
% \bottomrule
% \end{tabular}
% }
% \vspace{1em}
% \caption{\label{tab:base_model}Performance of different base models after training on the COIG Subset data.}
% \end{table}

\section{EXPERIMENTAL RESULTS}

\begin{table*}[t!]
\centering
\resizebox{0.95\textwidth}{!}{
\begin{tabular}{lccccccccccccc}
\toprule
\multirow{2}*{\textbf{Dataset}} & \multicolumn{7}{c}{\textbf{Instruction-Follow}} & \multicolumn{2}{c}{\textbf{Knowledge \& Reasoning}} & \multirow{2}*{\textbf{Average}} \\
\cmidrule(lr){2-8} \cmidrule(lr){9-10}
& \textbf{Gen.\& Sum.} & \textbf{Q\&A} & \textbf{CLS.} & \textbf{Rewrite} & \textbf{Extract} & \textbf{Math} & \textbf{Code} & \textbf{C-EVAL} & \textbf{CMMLU} &  \\
\midrule
\multicolumn{11}{c}{\textit{\textbf{Vanilla Model}}} \\
\midrule
Qwen2-7B & - & - & - & - & - & - & - & 83.2 & 83.9 & - \\
% Qwen2-72B & - & - & - & - & - & - & - &  &  & - \\
% LLaMA2-7B & - & - & - & - & - & - & - &  &  & - \\
% LLaMA2-13B & - & - & - & - & - & - & - &  &  & - \\
% LLaMA2-70B & - & - & - & - & - & - & - &  &  & - \\
\midrule
\multicolumn{11}{c}{\textit{\textbf{Qwen2-7B trained on baseline datasets}}} \\
\midrule
COIG & 68.60 & 48.05 & 39.5 & 59.0 & 46.8 & 42.9 & 28.6 & 69.5 & 72.0 & 62.0 \\
Firefly & 80.93 & 63.45 & \textbf{64.0} & \textbf{91.5} & 50.9 & 51.0 & 57.9 & 77.1 & 78.4 & 73.8 \\
Alpaca-zh & 79.77 & 56.30 & 57.5 & 88.0 & 63.6 & 15.2 & 40.7 & 77.7 & 77.3 & 69.9 \\
COIG-PC & 78.40 & 51.10 & 44.7 & 62.5 & 67.3 & 12.9 & 54.3 & 74.9 & 75.8 & 66.7 \\
OL-CC & \textbf{85.53} & 57.30 & 56.0 & 71.5 & 64.5 & 24.3 & 51.1 & 78.7 & \textbf{81.2} & 72.5 \\
\midrule
\multicolumn{11}{c}{\textit{\textbf{Qwen2-7B trained on \data }}} \\
\midrule
\data -Full & 77.40 & \textbf{65.80} & 55.0 & 59.0 & 64.5 & 66.7 & 65.7 & 76.9 & 77.5 & 73.1 \\
\data -Sub & 77.47 & 64.00 & 54.0 & 58.5 & \textbf{70.9} & \textbf{83.3} & \textbf{71.4} & \textbf{78.9} & 79.5 & \textbf{74.8} \\
% \midrule
% \multicolumn{11}{c}{\textit{Models with different parameter size trained on \data }} \\
% \midrule
% LLaMA2-7B & - & - & - & - & - & - & - &  &  & - \\
% LLaMA2-13B & - & - & - & - & - & - & - &  &  & - \\
% LLaMA2-70B & - & - & - & - & - & - & - &  &  & - \\
% Qwen2-72B & - & - & - & - & - & - & - &  &  & - \\
\bottomrule
\end{tabular}
}
\caption{Performance comparison with grouped Instruction-Follow tasks and detailed Knowledge \& Reasoning.}
\label{tab:baseline dataset}
\end{table*}

\vspace{-0.1cm}
\subsection{Ablating Instruction Data Sources and Base Models}
\vspace{-0.05cm}
We finetune the Qwen2-7B \citep{yang2024qwen2} and LLaMA2-13B \citep{touvron2023llama} models on different data sources from \data to analyze the impact of data sources on model capabilities across various domains. Then, we evaluate each model's performance on various types of assistant-style tasks using GPT-4o as LLM-as-Judge evaluator on Belle-Eval \citep{ji2023exploring}. Evaluation details are provided in Appendix \ref{app:detail eval}.
% To understand the correlation between training data sources and the downstream performance of different tasks, we evaluate the models on 10 tasks from BELLE-Eval. We employ GPT-4 as evaluator for scoring model responses, with scores ranging from 0 to 1.

% To understand the correlation between training data sources and the downstream performance of different tasks, we evaluate the models on 10 tasks from BELLE-Eval. We employ GPT-4 as evaluator for scoring model responses, with scores ranging from 0 to 1.
Tabel \ref{tab:belle} shows the performance of Qwen2-7B and LLaMA2-13B models fine-tuned on different subsets. The table indicates that all fine-tuned models achieved significant improvements across various domains. Notably, the Qwen model trained on the Ruozhiba dataset performed remarkably well, even surpassing high-quality data subsets like COIG-PC and Zhihu. Despite the fact that Ruozhiba is not commonly recognized in the Chinese academic community and often contains humorous or absurd content, we believe these characteristics contributed to its effectiveness. 
The Ruozhiba dataset has inherent logical structures, includes cognitive and linguistic traps, and features jokes and riddles, as well as artistic and abstract rhetorical techniques. These elements, in turn, challenge the model’s multi-hop reasoning capabilities, enhancing its understanding of the Chinese language during fine-tuning and improving its capacity for complex logical reasoning.
Human Value ranks second on average across all subsets, which aligns with expectations, as this subset contains a substantial amount of high-quality human-annotated data that aligns well with human values. This data not only improved instruction-following capabilities during fine-tuning but also prevented models from biasing towards specific values, enhancing universality.
Moreover, WikiHow scores only 30.2 on Qwen and 18.6 on LLaMA-2-13B, likely due to the lack of diversity in its "how-to" instructions.

\begin{table}[t]
\centering
\resizebox{0.85\columnwidth}{!}{
\begin{tabular}{lccc}
\toprule
\textbf{Source} & \textbf{I.F.} & \textbf{K\&R} & \textbf{Average} \\
\midrule
\multicolumn{4}{c}{\textit{Single Domain}} \\
\midrule
NLP & 63.0 & 76.7 & 69.9 \\
Exam & \textbf{70.6} & 80.8 & \textbf{75.7} \\
Wiki & 59.2 & 83.1 & 71.2 \\
Social\&Forum & 67.3 & \textbf{83.2} & 75.3 \\
\midrule
\multicolumn{4}{c}{\textit{Mixed Domain}} \\
\midrule
NLP+Wiki & 57.5 & 68.9 & 63.2 \\
NLP+Exam & 65.2 & 80.7 & 73.0 \\
NLP+Social\&Forum & 65.6 & 78.9 & 72.3 \\
Exam+Wiki & 69.0 & 79.6 & 74.3 \\
Exam+Social\&Forum & \textbf{70.4} & \textbf{81.1} & \textbf{75.8} \\
\bottomrule
\end{tabular}}
\caption{Comparison of Data Mixing Strategy on Instruction-Follow, Knowledge\&Reasoning.}
\label{tab:training strategy}
\end{table}

\begin{table*}[t!]
\centering
\resizebox{0.96\textwidth}{!}{%
\begin{tabular}{lcccccc}
\toprule
\textbf{Model}  & \textbf{BBH (3-Shot, CoT)} & \textbf{GSM8k (4-shot)} & \textbf{HumanEval (P@10)} & \textbf{TydiQA (GP, 1-shot)} & \textbf{Average} \\ 
\midrule
\multicolumn{6}{c}{\textit{\textbf{Official Models}}} \\
\midrule
Yi-6B    & 45.7  & 34.5  & 28.3  & 47.6   & 39.0  \\ 
Yi-6B-Chat & 43.0  & 38.0  & 31.4  & 24.8  & 34.3  \\ 
Qwen2-7B & 59.5  & 71.5  & 76.0  & 62.3  & 67.3  \\
Qwen2-7B-Instruct & 63.7  & 84.5  & 87.6  & 34.5  & 67.5  \\ 
LLaMA-2-7B & 41.6  & 14.5  & 25.2  & 43.5   & 31.2  \\ 
LLaMA-2-7B-Instruct & 21.7  & 8.5   & 25.2  & 20.9  & 19.1  \\ 
\midrule
\multicolumn{6}{c}{\textit{\textbf{Models trained on Open-Sourced Data Mixture}}} \\
\midrule
Yi-6B & 43.6 (\textcolor{persianred}{-2.1}) & 33.0 (\textcolor{persianred}{-1.5}) & 41.1 (\textcolor{oceanboatblue}{+12.8}) & 33.6 (\textcolor{persianred}{-14.0}) & 37.8 (\textcolor{persianred}{-1.2}/\textcolor{oceanboatblue}{+3.5}) \\ 
Qwen2-7B & 56.9 (\textcolor{persianred}{-2.6}) & 73.2 (\textcolor{oceanboatblue}{+1.7}) & 79.1 (\textcolor{oceanboatblue}{+3.1}) & 49.2 (\textcolor{persianred}{-13.1}) & 64.6 (\textcolor{persianred}{-2.7}/\textcolor{persianred}{-2.9}) \\ 
LLaMA-2-7B & 40.1 (\textcolor{persianred}{-1.5}) & 27.0 (\textcolor{oceanboatblue}{+12.5}) & 37.0 (\textcolor{oceanboatblue}{+11.8}) & 23.1 (\textcolor{persianred}{-23.4}) & 31.8 (\textcolor{oceanboatblue}{+0.6}/\textcolor{oceanboatblue}{+12.7}) \\ 
\midrule
\multicolumn{6}{c}{\textit{\textbf{Models trained on \data}}} \\
\midrule
Yi-6B & 43.7 (\textcolor{persianred}{-2.0}) & 21.4 (\textcolor{persianred}{-13.1}) & 25.9 (\textcolor{persianred}{-2.4}) & 50.0 (\textcolor{oceanboatblue}{+2.4}) & 34.9 (\textcolor{persianred}{-4.1}/\textcolor{oceanboatblue}{+0.6}) \\ 
Qwen2-7B & 60.7 (\textcolor{oceanboatblue}{+1.2}) & 77.1 (\textcolor{oceanboatblue}{+5.6}) & 80.3 (\textcolor{oceanboatblue}{+4.3}) & 63.3 (\textcolor{oceanboatblue}{+1.0}) & 70.4 (\textcolor{oceanboatblue}{+3.1}/\textcolor{oceanboatblue}{+2.9}) \\ 
LLaMA-2-7B & 39.7 (\textcolor{persianred}{-1.9}) & 13.2 (\textcolor{persianred}{-1.3}) & 24.6 (\textcolor{persianred}{-0.6}) & 49.1 (\textcolor{oceanboatblue}{+5.6}) & 31.7 (\textcolor{oceanboatblue}{+0.5}/\textcolor{oceanboatblue}{+12.6}) \\ 
\midrule
\multicolumn{6}{c}{\textit{\textbf{Models trained on \data\ + Open-Sourced Data Mixture}}} \\
\midrule
Yi-6B & 44.9 (\textcolor{persianred}{-0.8}) & 29.5 (\textcolor{persianred}{-5.0}) & 40.8 (\textcolor{oceanboatblue}{+12.5}) & 37.6 (\textcolor{persianred}{-10.0}) & 38.2 (\textcolor{persianred}{-0.8}/\textcolor{oceanboatblue}{+3.9}) \\ 
Qwen2-7B & 60.9 (\textcolor{oceanboatblue}{+1.4}) & 75.4 (\textcolor{oceanboatblue}{+3.9}) & 80.9 (\textcolor{oceanboatblue}{+4.9}) & 64.2 (\textcolor{oceanboatblue}{+1.9}) & 70.4 (\textcolor{oceanboatblue}{+3.1}/\textcolor{oceanboatblue}{+2.9}) \\ 
LLaMA-2-7B & 41.7 (\textcolor{oceanboatblue}{+0.1}) & 26.5 (\textcolor{oceanboatblue}{+12.0}) & 38.6 (\textcolor{oceanboatblue}{+13.4}) & 45.3 (\textcolor{oceanboatblue}{+1.8}) & 38.0 (\textcolor{oceanboatblue}{+6.8}/\textcolor{oceanboatblue}{+18.9}) \\ 
\bottomrule
\end{tabular}
}
\caption{Performance comparison of different models and versions on various non-Chinese tasks. Numbers in parentheses represent differences from the base model. For the averages, the left value represents the difference from the base model, and the right value represents the difference from the chat model.}
\label{tab:open ins}
\end{table*}

% % 
We also evaluated different base models with varying parameter sizes fine-tuned on the \data-Sub. The table \ref{tab:different base model} presents the performance differences across models in instruction-following and knowledge \& reasoning tasks. As expected, model size correlates with improved performance across all tasks. The Yi and Qwen2 series show strong results, with Qwen2-72B leading overall. LLaMA-2 series lags behind as it wasn’t specifically designed for Chinese language understanding. 
Additionally, we assess the safety performance of various fine-tuned models on SafetyBench \citep{zhang2023safetybench}. The detailed experimental results can be found in Appendix \ref{app:safe}.

\subsection{Comparison with Other Chinese Instruct-tuning Datesets}

Table \ref{tab:baseline dataset} illustrates how models trained on different datasets perform various instruction-following tasks. \data stands out in Q\&A, extraction, math, and coding tasks, indicating its strength in knowledge-intensive and reasoning areas. However, for classification, summarization, and rewrite tasks (e.g., translation and text editing), \data underperforms. Our case study attributes this to the limited representation of these tasks in the dataset. To improve performance in these areas, we recommend augmenting with datasets such as Firefly and Alpaca-Zh.
% Additionally, we visualized the semantic distribution of these datasets using U-MAP, as shown in Appendix \ref{}, which reveals that \data exhibits the broadest semantic distribution among the compared datasets.

\subsection{Exploration of Data Mixture Strategy}
\subsubsection{Mixture of Different Domain}

We categorized \data into four main sources as described earlier: NLP datasets, Exams, World Knowledge, and Social Media \& Forums. We evaluated different combinations of these sources across tasks, as shown in Table \ref{tab:training strategy}.
Our results show that data from Social \& Forum and Exam sources most significantly enhance the model’s instruction-following ability. World Knowledge and Social \& Forum data, meanwhile, contribute to improved knowledge performance, aligning with expectations: Social \& Forum and Exam data cover broader, more complex tasks, while NLP and World Knowledge data tend to focus on more constrained, traditional tasks. World Knowledge naturally excels in knowledge-intensive tasks such as C-Eval and CMMLU.

When mixing data sources, we observed that combinations of two domains rarely outperform the stronger individual source. Mixing weaker sources, such as NLP and Wiki, can result in further degradation, especially on instruction-following tasks. However, combining two strong sources tends to maintain high performance, such as mixing Exam and Social \& Forum data for instruction-following, or Wiki with Social \& Forum for knowledge and reasoning tasks.

\begin{table}[t]
\centering
\resizebox{0.83\columnwidth}{!}{
\begin{tabular}{lcccc}
\toprule
\textbf{Model Series} & \textbf{Size} & \textbf{I.F.} & \textbf{K\&R} & \textbf{Average} \\
% \midrule
% \multicolumn{5}{c}{\textit{Vanilla Models}} \\
% \midrule
\midrule
\multirow{2}*{Yi} & 6B & 55.5 & 74.1 & 64.8 \\
& 34B & 62.3 & 77.6 & 70.0 \\
\midrule
\multirow{3}*{LLaMA-2} & 7B & 35.6 & 33.2 & 34.4 \\
 & 13B & 43.5 & 38.0 & 40.8 \\
 & 70B & 47.7 & 50.2 & 48.9 \\
 \midrule
\multirow{2}*{Qwen2} & 7B & 70.3 & 79.2 & 74.8 \\
& 72B & 73.3 & 89.8 & 81.6 \\
\bottomrule
\end{tabular}
}
\caption{Model performance comparison across different model series and sizes on Instruction-Following and Knowledge \& Reasoning tasks.}
\label{tab:different base model}
\end{table}

\subsubsection{Mixed with Open-Sourced data}\label{sec:os mix}

To explore \data’s potential on non-Chinese tasks, we extended our evaluation using the open-instruct suite\citep{wang2023far} to include four additional non-Chinese tasks: BBH (reasoning), GSM8K (math), HumanEval (code), and TydiQA (multilingualism). The results are shown in the table \ref{tab:open ins}.

Models trained on \data performs on par with official base and chat models on BBH and significantly outperform baselines on TydiQA, highlighting its strength in activating multilingual capabilities. Given that \data is a Chinese-focused dataset with less than 5\% of math and code data, it understandably underperforms on English-heavy tasks like GSM8K and HumanEval, particularly when using Yi-6B as the base model. The use of purely Chinese data impacts performance in these tasks. In contrast, experiments with Qwen2-7B and Llama2-7B show \data performing at or above the level of base models, although it still lags behind the more data-engineered chat models. This gap is expected, as these official chat models undergo extensive, costly data engineering.

To address the language mismatch, we experimented with mixing \data with open-source English datasets. We sampled equivalent data from Magicoder (code) and Mammoth (math), labeled as OS Mix, and combined it with CQIA. Our findings show:
(1) \data matches or exceeds OS Mix on reasoning tasks and performs significantly better on multilingual tasks.
(2) On code and math tasks, \data’s performance varies with the base model. With Yi-6B and Llama2-7B, \data lags behind OS Mix, which is specialized for these tasks. Surprisingly, using Qwen2-7B as the base, \data outperforms OS Mix on the same tasks.
(3) Combining \data with OS Mix strengthens each dataset’s weaknesses, leading to overall performance gains.

\begin{figure}[tp]
    \centering
    \includegraphics[scale=0.32]{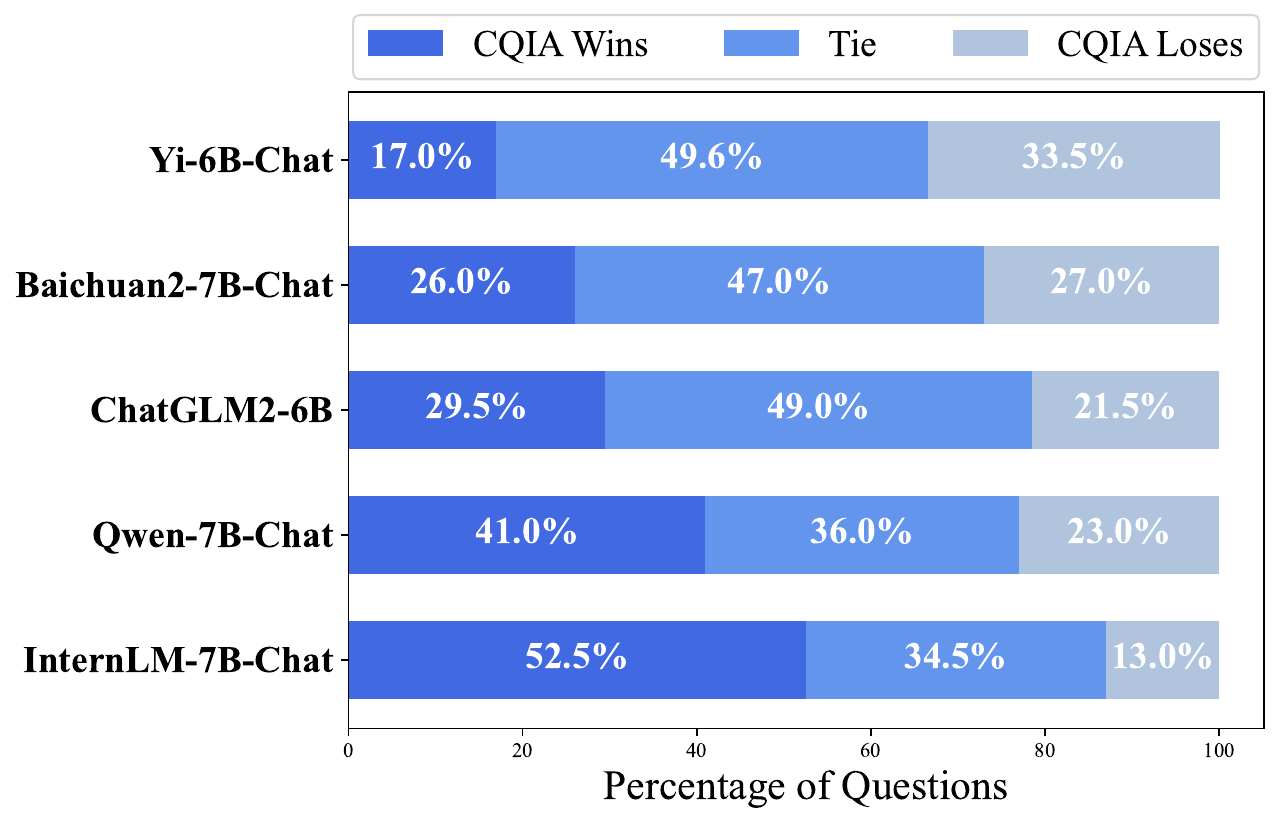}
    \caption{\label{fig:human eval}Human evaluation of pair-wise comparison between \data-Sub and 5 strong baselines.}
\end{figure}
\subsection{Human Evaluation}

We compared Yi-6B\citep{young2024yi} fine-tuned on the \data-Sub with several Chinese open-source chat models. Focusing on real-world questions, we sampled 200 prompts from OL-CC\footnote{https://data.baai.ac.cn/details/OL-CC} and Zhihu, none of which were part of the training set. We conducted a pairwise comparison to assess how our model performs in real-world scenarios. Figure \ref{fig:human eval} presents the human evaluation results comparing \data against five baselines: Yi-6B-Chat \citep{young2024yi}, Baichuan2-7B-Chat \citep{baichuan2023baichuan2}, ChatGLM2-6B \citep{glm2024chatglm}, Qwen-7B-Chat \citep{bai2023qwen}, and InternLM-7B-Chat \citep{cai2024internlm2}. The results show that the model trained on \data achieved higher human preference, with over 60\% responses being rated as better or tied with the baselines. This demonstrates \data’s ability to align more closely with real-world human communication patterns, resulting in higher user preference. See details in Appendix \ref{app:detail eval}.

% \vspace{-0.3cm}
\subsection{Data Scaling}
\begin{figure}[t!]
    \centering
    \includegraphics[scale=0.23]{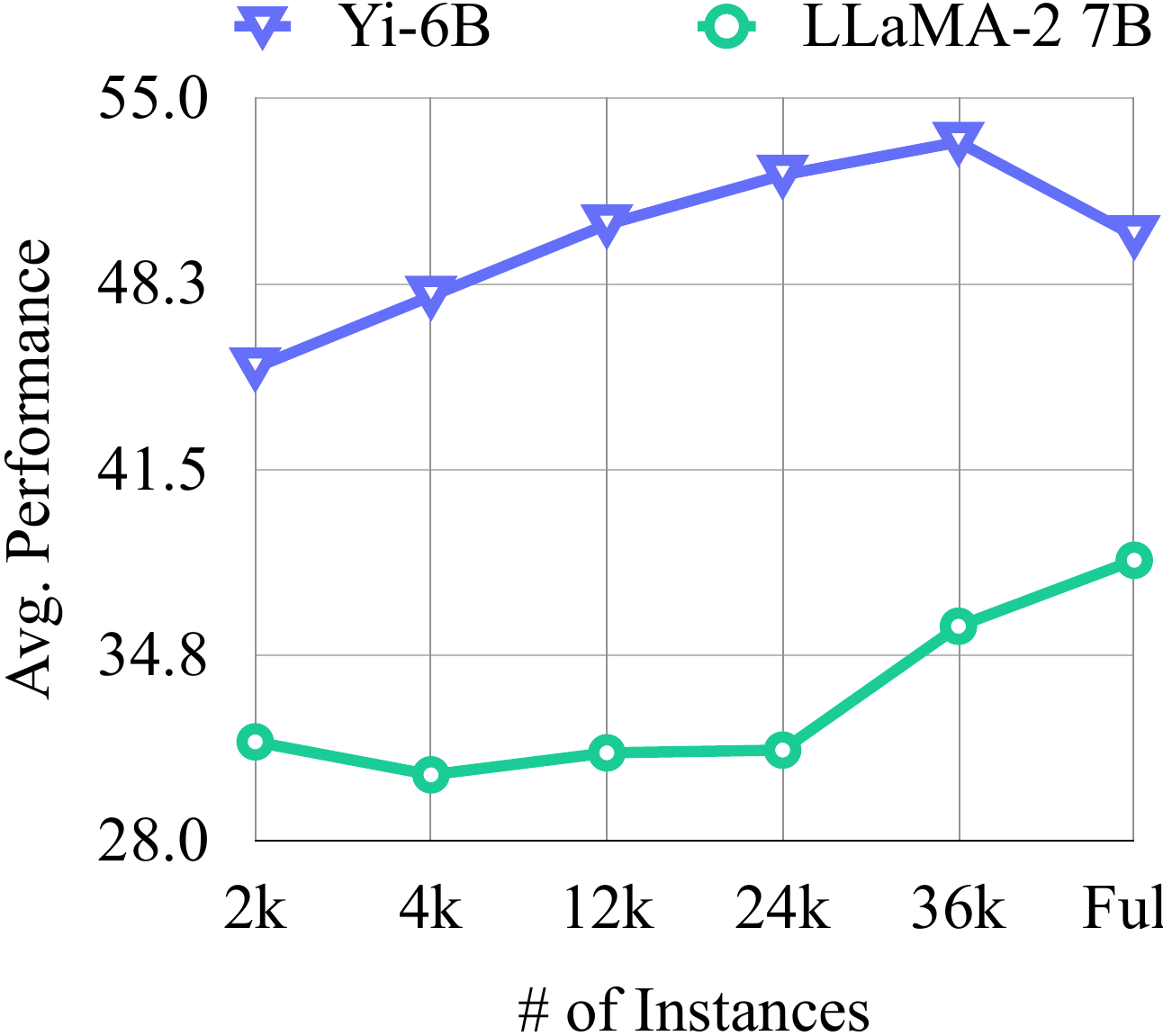}
    \caption{Performance of the Model Trained on Different Data Scales.}
    \label{fig:data_scaling}
\end{figure}
The data scaling results in Figure \ref{fig:data_scaling} demonstrate the impact of training set size on model performance for Yi-6B and LLaMA-2-7B. Both models exhibit performance improvements as the number of instances increases, underscoring the significance of data quantity in enhancing language model capabilities. Yi-6B shows rapid gains up to 24k instances, after which performance stabilizes with minor fluctuations. This plateau effect may be attributed to the limitations of our dataset scale, where model behavior becomes less predictable at this data size scale. In contrast, LLaMA-2 7B displays a consistent upward trend across the entire range.

\section{Related Work}

\subsection{Instruction-Tuning Dataset}
% Instruction tuning enhances large language models (LLMs) by training them to generate responses aligned with input instructions, enabling conversational and task execution capabilities. 
Instruction tuning enhances the conversational and task execution capabilities of large language models (LLMs) by training them to generate responses aligned with input instructions.
This approach yields more controllable and predictable models that better align with human intent. Several strategies have been employed to construct instruction-tuning datasets: (1) Manual annotation by human experts  \citep{conover2023free}. (2) Repurposing existing NLP datasets  \citep{mishra2022crosstask, sanh2022multitask, chung2022scaling}. (3) Synthesising using LLMs \citep{honovich2022unnatural, wang2023selfinstruct, xu2023wizardlm, ji2023exploring, xu2023baize}. While efficient, this method may introduce inconsistencies and noise.
While numerous English instruction tuning datasets exist, their Chinese counterparts are limited. Some efforts focus on translating English datasets \citep{peng2023instruction}, while others repurposing existing NLP tasks into instruction formats  \citep{coig-pc, Firefly}. Notable Chinese datasets include HC3  \citep{guo2023close}, COIG  \citep{zhang2023chinese}, BELLE  \citep{ji2023exploring}, and MOSS  \citep{sun2023moss}.

\subsection{Data Mixture Strategies for SFT}
% \vspace{-0.2cm}
Recent research emphasizes the importance of data quality in instruction tuning. LIMA  \citep{zhou2023lima} demonstrates strong performance using only 1,000 high-quality instruction-output pairs. AlpaGasus  \citep{chen2023alpagasus} and Humpback  \citep{li2023selfalignment} employ advanced filtering techniques to enhance dataset quality and training efficiency.
Studies also explore the impact of mixing different instruction-tuning datasets. \citet{song2023dynamics} investigate various combination approaches, while the Tulu series  \citep{konchakov2023critical, ivison2023camels} demonstrates that increasing instruction diversity can improve overall performance. Notably, no single dataset or combination consistently outperforms others across all metrics, highlighting the complexity of optimizing instruction-tuning data mixtures.

% Currently, more and more studies has begun to pay attention to the importance of data quality of instruction tuning. LIMA \citep{zhou2023lima} only uses 1,000 high-quality instructions and outputs for instruction-tuning and does not even need to perform RLHF training to achieve very strong performance. AlpaGasus \citep{chen2023alpagasus} uses powerful LLM to automatically identify and filter low-quality data, resulting in high-quality instruction tuning data to improve performance and training speed. Humpback \citep{li2023selfalignment} filters out high-quality samples to fine-tune a more powerful LLM.

% Others \citep{song2023dynamics} explores the impact of the mixture strategies of different instruction tuning datasets. Tulu series \citep{konchakov2023critical, ivison2023camels} show that increasing instruction diversity can effectively improve the performance and different instruction tuning datasets can discover or enhance specific skills, while no one dataset (or combination) provides the best performance across all assessments.

\section{Conclusion}
%In this paper, we introduce a a high-quality Chinese instruction fine-tuning dataset. COIG-CQIA focuses on creating a dataset from various website from Chinese internet. These are deeply cleansed, restructured, and manually reviewed to ensure quality, diversity, and relevance. This dataset is designed to provide the Chinese NLP community with high-quality and human interaction-aligned instruction-tuning data. 
This paper presents the COIG-CQIA, a high-quality Chinese instruction fine-tuning dataset. The dataset is compiled from various websites on the Chinese internet, then meticulously cleansed, restructured, and manually reviewed to guarantee its quality, diversity, and relevance. It aims to offer the Chinese NLP community with meticulously crafted fine-tuning data that aligns with human interaction.

\section{Limitation}

We acknowledge several limitations in our study. While \data is comprehensive, the inclusion of subjective elements may lead to varying interpretations, potentially impacting data construction. Additionally, our focus on Chinese language data covers only a fraction of human knowledge. The evaluation metrics may not fully capture the models’ sophisticated understanding and reasoning abilities. These limitations underscore the need for ongoing refinement and expansion of our dataset. In future work, we aim to collect and aggregate more diverse Chinese instruction-tuning data to improve the models’ capability and reliability.

\section{Ethics Statement}

In developing \data, we strictly adhere to ethical guidelines and legal regulations, ensuring fairness, transparency, inclusivity and respect for all stakeholders. We stress the importance of safeguarding privacy and intellectual property rights, underscoring our commitment to responsible and lawful data management. We have taken steps to anonymize any personal data to protect privacy and have made every effort to minimize harmful or biased content. However, we recognize that biases can inadvertently arise and some information may be potentially offensive. We are committed to continuous monitoring and improvement to mitigate such biases. Furthermore, we encourage users of our dataset to employ it responsibly and to consider the ethical implications of their work, particularly in applications that may impact individuals or communities.

% \section*{Acknowledgments}

\bibliography{ARR_BIB}

\clearpage
\appendix
\section{Details of COIG-CQIA Curation}\label{app:data curation}

\subsection{Social Media \& Forums}

We curated data from five prominent Chinese social media platforms and forums, each offering unique content characteristics.

\paragraph{Zhihu} \label{appdix:Zhihu}is a comprehensive Q\&A platform where users can ask and answer questions on various topics, making it an extensive repository of knowledge and insights. However, the absence of a review mechanism for answers on Zhihu leads to a large volume of content that falls short of our quality standards. 
To address this issue, we implemented a multi-step filtering process.
%\fft{short version: First, we selected answers with over 50 upvotes, reducing the dataset from 10 million to 2 million entries. Next, we applied a rule-based filter to remove content with sensitive keywords, narrowing it to 100K. Then, GPT-4 evaluated the remaining responses, retaining only those scoring above 8, resulting in 8K high-quality answers. Finally, human annotators reviewed and selected the top 5.6K entries for our final dataset.}
Initially, we selected answers that had garnered more than 50 upvotes, which reduced our original dataset from 10 million entries to 2 million. We then applied a rule-based method to filter out content containing sensitive or potentially harmful keywords, further narrowing the dataset to 100K. Subsequently, we leveraged GPT-4 to evaluate the remaining responses on a scale of 1-10, retaining only those that scored above 8, which yield approximately 8K high-quality answers. In the final step, human annotators carefully reviewed and selected the top 5.6K entries, ensuring that only the highest quality, most informative content was included in our final dataset. 
% To filter low quality answers, we selected answers with more than 50 upvotes, then filtering out content containing sensitive or harmful keywords using a rule-based method. Subsequently, we employed GPT-4 to score the responses on a scale of 1-10, retaining those with scores above 8.

\paragraph{SegmentFault} \label{appdix:SegFault}is IT-focused Q\&A community which is similar to Stack Overflow in its scope and purpose. 
To ensure the relevance and currency of the dataset, we concentrated on content posted after 2018, acknowledging that earlier posts might be outdated due to evolving programming languages and software versions. Our selection process prioritized "accepted" answers that had received a minimum of 5 upvotes, indicating community validation of their quality and usefulness. To further refine our dataset, we conducted a comprehensive manual review of all instruction-response pairs. 
% Our data are collected from the contents posted before 2018, as earlier content may become outdated due to changes in programming languages or software versions. We then select the "accepted" answers with at least 5 upvotes. Furthermore, we manually review all the (question, answer) pairs to remove or modify low-quality content.

\paragraph{Douban} \label{appdix:Douban}is a social platform focused on literature and arts, where users share content related to books, movies, TV series, music, and more. We sampled data from books, movies, and TV series, collecting metadata such as ratings, actor/crew details, and long reviews. 
Based on this rich dataset, we created three main tasks: synopsis generation, review generation, and recommendations.
% Then, we design three tasks in total: synopsis generation, review generation, and recommendations.
% For each task, we manually design various prompt templates and used these templates in combination with metadata to construct instructions. 
For each task, we designed a variety of prompt templates, combining them with metadata to construct comprehensive instructions. In the case of synopsis and review generation, we utilized movie or TV series names in conjunction with these templates, using Douban user-generated content as responses. We then applied quality filtering to elimminate short or irrelevant answers and remove personal information.
% For synopsis generation and review generation, we construct instructions using prompt templates combined with movie or TV series names, with responses generated by Douban users. Then we remove responses with lengths shorter than a threshold and delete personal information and irrelevant content(e.g., "Subscribe our Official Accounts"). 
To improve real-world applicability, we refined some instructions to include more implicit intents, aligning responses more closely with the content.

\paragraph{Xiaohongshu} \label{appdix:Xiaohongshu}is a popular social media platform in China that serves as a hub for users to share their daily lives, travel experiences, food, and product recommendations. Contents in this platform are renowned on the Chinese internet for their unique expressive style. For our study, we curated a sample of posts ranging from 500 to 2000 characters in length. To maintain focus on the core content, we excluded posts that contained user interactions (such as "@User\_Name" mentions) or references to visual media (e.g., "as shown in the picture/video"). 

\paragraph{Ruozhiba} \label{appdix:Ruozhiba}is a sub-forum within Baidu Tieba, China's largest interest-based online community platform. 
This particular forum is renowned for its linguistic complexity, featuring posts rich in wordplay, including puns, polysemous terms, causal reversals, and homophones. Many of these posts are ingeniously crafted with logical traps that present cognitive challenges even for native speakers. 
% Its posts often contain puns, polysemous terms, causal reversals, and homophones, many of which are designed with logical traps, posing challenges even for humans. 
In our study, we focus on the 500 most upvoted threads in this forum. We used the thread titles as potential instructions, carefully filtering out those that were non-instructive (such as mere declarative statements or unanswerable queries) or contained toxic content. For answer curation, human evaluators first identified the traps within the instructions and then prompted GPT-4 to generate responses. This process was repeated until GPT-4 produced correct answers.
% We collected the 500 most upvoted threads. Using the titles as instructions, we eliminate those that were either non-instructive (i.e., declarative statements or unanswerable) or toxic.  Responses were generated by either humans or GPT-4. We conducted manual reviews for GPT4's responses to ensure accuracy, ultimately obtaining 240 (instruction, response) pairs.

\subsection{World Knowledge}
Introducing world knowledge to LLMs is crucial for enhancing their ability to engage in knowledge-driven interactions. To collect comprehensive data in this broad field of information, we focused on two key areas.

\subsubsection{General Encyclopedia} \label{appdix:GeneralEncyclopedia}
General encyclopedias provide comprehensive coverage of a wide range of topics across various fields. We collected data from three prominent Chinese encyclopedic websites: One Hundred Thousand Whys, wikiHow-zh, and Encyclopedia of China.
\textbf{One Hundred Thousand Whys} focuses on popular science, featuring articles that ask "why" across diverse topics. We collected data from all 15 categories, using article titles as instructions and content as responses, filtering out responses under 300 characters.
\textbf{WikiHow-zh}, \label{appdix:wikihow}the Chinese version of WikiHow, covers a wide range of "how-to" articles. We sampled 1.5K entries from all 19 categories, filtered for quality and length, and used titles as instructions and article contents as responses.
\textbf{Encyclopedia of China} is a comprehensive resource with 500K expert-authored entries. We designed various prompt templates for concept explanation tasks, sampling entries from all 74 categories. Instructions were constructed by combining entry names or subtitles with prompt templates, with corresponding content used as responses.

\subsubsection{Domain Specific Knowledge}

We collected data from four specific domains: medicine, economic management, electronics, and agriculture.
\noindent \textbf{Medical Domain} data was sourced from three websites: Baobaozhidao, Qianwen Health, and Baikemingyi. The first two feature expert-written Q\&A articles, while Baikemingyi offers structured data on diseases and medications. We used article titles as instructions and content as responses, designing various prompt templates for structured data.
\noindent \textbf{Economic Management Domain} data came from MBA Wiki Encyclopedia, a collaborative knowledge platform. We created instructions by combining entry names with designed prompt templates, using entry content as responses.
\noindent \textbf{Electronics Domain} data was collected from the EETrees electronic encyclopedia, following a similar method of combining entry names with prompt templates to create instruction-response pairs.
\noindent \textbf{Agriculture Domain} data was sourced from an agricultural encyclopedia website covering various topics. We constructed instruction-response pairs from article titles and content, applying specific filtering criteria.
% We collected data from four specific domains: medicine, economic management, electronics, and agriculture.

\subsection{Examinations}
To equip the model with robust problem-solving skills and a comprehensive knowledge foundation, we leveraged a diverse range of examination resources in the training process. 

\paragraph{The Middle School and College Entrance Examinations}\label{appdix:mid_school_exam} data is primarily sourced from the COIG dataset\cite{zhang2023chinese}, focusing on China's principal general competency tests. These data cover various humanities subjects and include detailed answer explanations. After filtering and processing, we obtained 1964 (instruction, response) pairs. 
% Details on data processing and subject coverage are provided in Appendix C.1\ref{}.
\paragraph{Graduate Entrance Examination}\label{appdix:Graduate Entrance Exam} data represents one of China's most challenging assessments, covering advanced knowledge across multiple disciplines. We collected and processed recent exam papers, converting them to LaTeX format and verifying their accuracy. 
% Specific disciplines and data processing methods are outlined in Appendix C.2\ref{}.
\paragraph{Logical Reasoning Test} \label{appdix:Logical exam}data aims to assess critical thinking and problem-solving skills. We collected logic reasoning questions with detailed answer analyses from various online sources. 
% See Appendix C.3\ref{} for data collection criteria.
\paragraph{Chinese Culture Test} \label{appdix:chinese culture}data investigates the mastery of traditional Chinese culture and history. We compiled multiple-choice questions with answer analyses from online resources. 
% Data collection and processing details are available in Appendix C.4\ref{}.

\subsection{NLP Datasets}
To further enhance the model's language understanding and generation capabilities, we incorporated several specialized NLP datasets into \data. These datasets were carefully selected to cover a wide range of linguistic tasks and cultural contexts.

\paragraph{COIG-PC}\label{appdix:COIG_PC} is a comprehensive collection of Chinese NLP tasks. We selected and processed a subset of these tasks, focusing solely on Chinese-language data for information extraction, classification, and summarization, etc. After sampling and human verification, we obtained 3K high-quality instruction-response pairs. Notably, while this dataset provides valuable task-specific training, its characteristically short outputs required careful integration to avoid compromising the model's overall chat performance across tasks. 

\paragraph{COIG Human Value}\label{appdix:COIG_Human_value}, a subset of the COIG dataset\cite{zhang2023chinese}, focuses on instruction fine-tuning data aligned with Chinese cultural values. We filtered and processed this data to ensure quality and relevance.

\begin{table*}[htbp]
\centering
\begin{tabular}{lrlcc}
\toprule
\textbf{Dataset} & \textbf{\# Instances} & \textbf{Source} & \textbf{Human Generated?} & \textbf{Human Verified?} \\
\hline
COIG & 178k & Existing Dataset\&Synthesis & $\times$ & $\times$ \\
Firefly & 1.1M & Existing Dataset & $\times$ & $\times$ \\
Alpaca-zh & 51k & Synthesis & $\times$ & $\times$ \\
COIG-PC & 321M & Existing Dataset & $\times$ & $\times$ \\
OL-CC & 10k & Human\&Synthesis & $\checkmark$ & $\times$ \\
\midrule
\data & 44k & Human & $\checkmark$ & $\checkmark$ \\
\bottomrule
\end{tabular}
\caption{Comparison of Different Datasets.}
\label{tab:dataset-comparison}\label{tab:comparison dataset}
\end{table*}

\paragraph{Firefly Chinese Traditional} \label{appdix:Firefly}comprises three tasks related to traditional Chinese culture: Classical Chinese Translation, Ancient Poetry Writing, and Idiom Interpretation. We sampled and filtered data from each task to ensure quality and appropriate length.

\paragraph{CValue} \label{appdix:CValue} addresses anti-discrimination and empathy across various dimensions. It includes human-generated prompts and expert-crafted responses aligned with human values. We incorporated all data from CValue to enhance the model's alignment with ethical considerations.

\begin{figure*}[t!]
    \centering
    \begin{minipage}{0.45\textwidth}
        \centering
        \includegraphics[width=\textwidth]{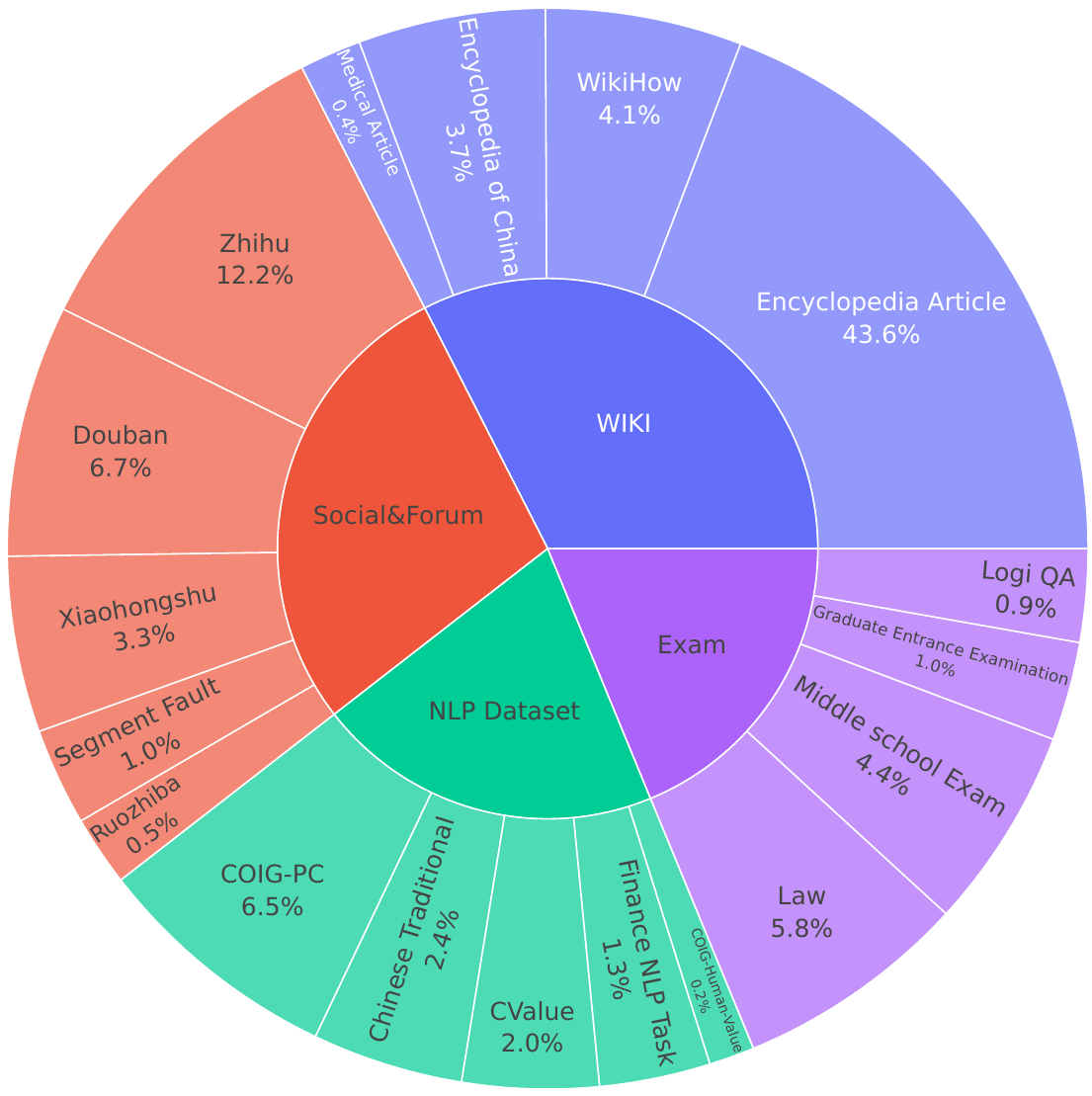}
        \caption{Data distribution.}
        \label{fig:source distribution}
    \end{minipage}
    % \hfill
    \hspace{0.02\textwidth}
    \begin{minipage}{0.45\textwidth}
        \centering
        \includegraphics[width=\textwidth]{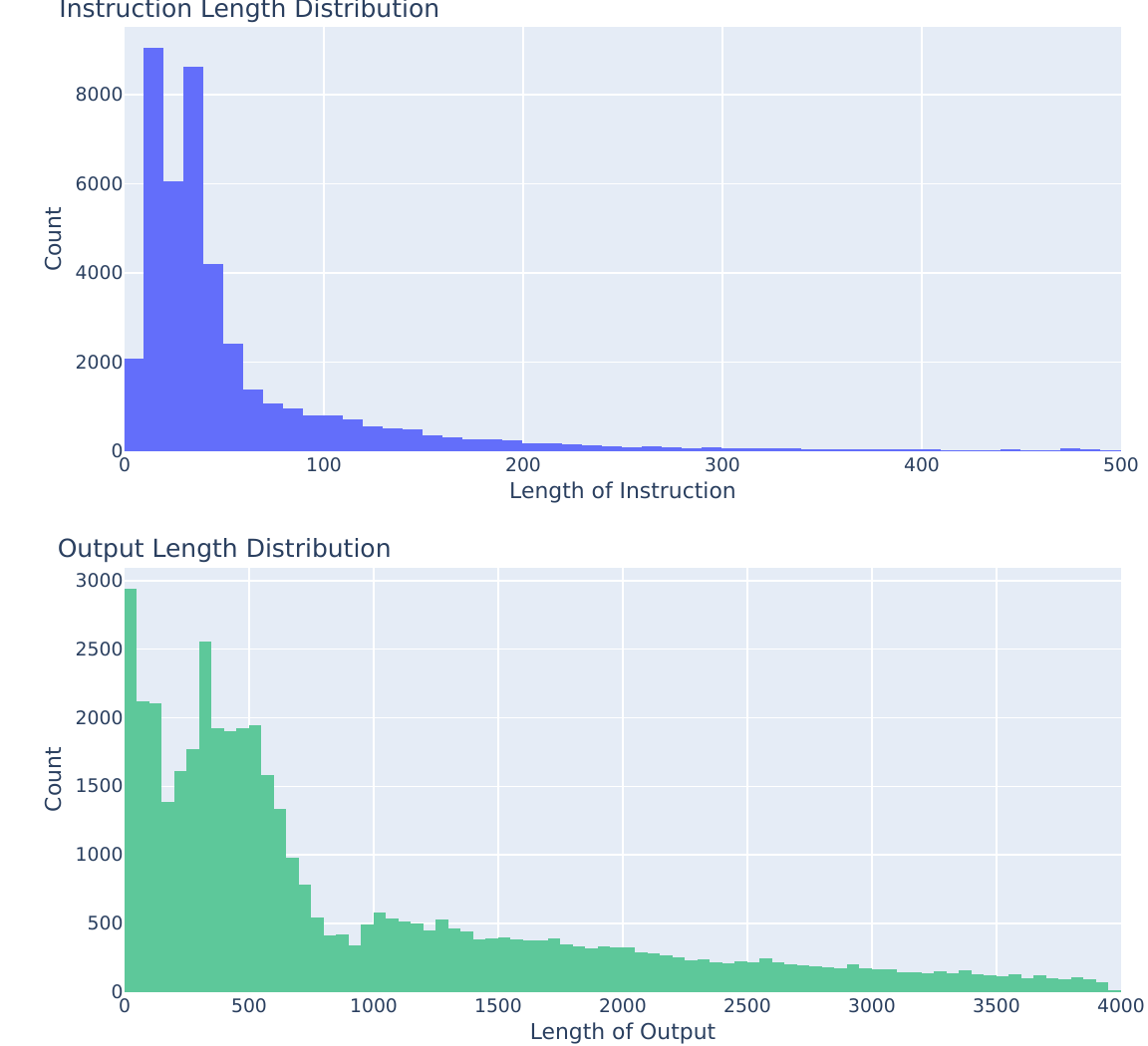}
        \caption{Length distribution of instruction and responses. Note that the instruction is the concatenation of instructions and inputs in \data.}
        \label{fig:sequence_length}
    \end{minipage}
\end{figure*}
% \clearpage

\begin{figure*}[htbp]
\centering
\includegraphics[width=0.75\textwidth]{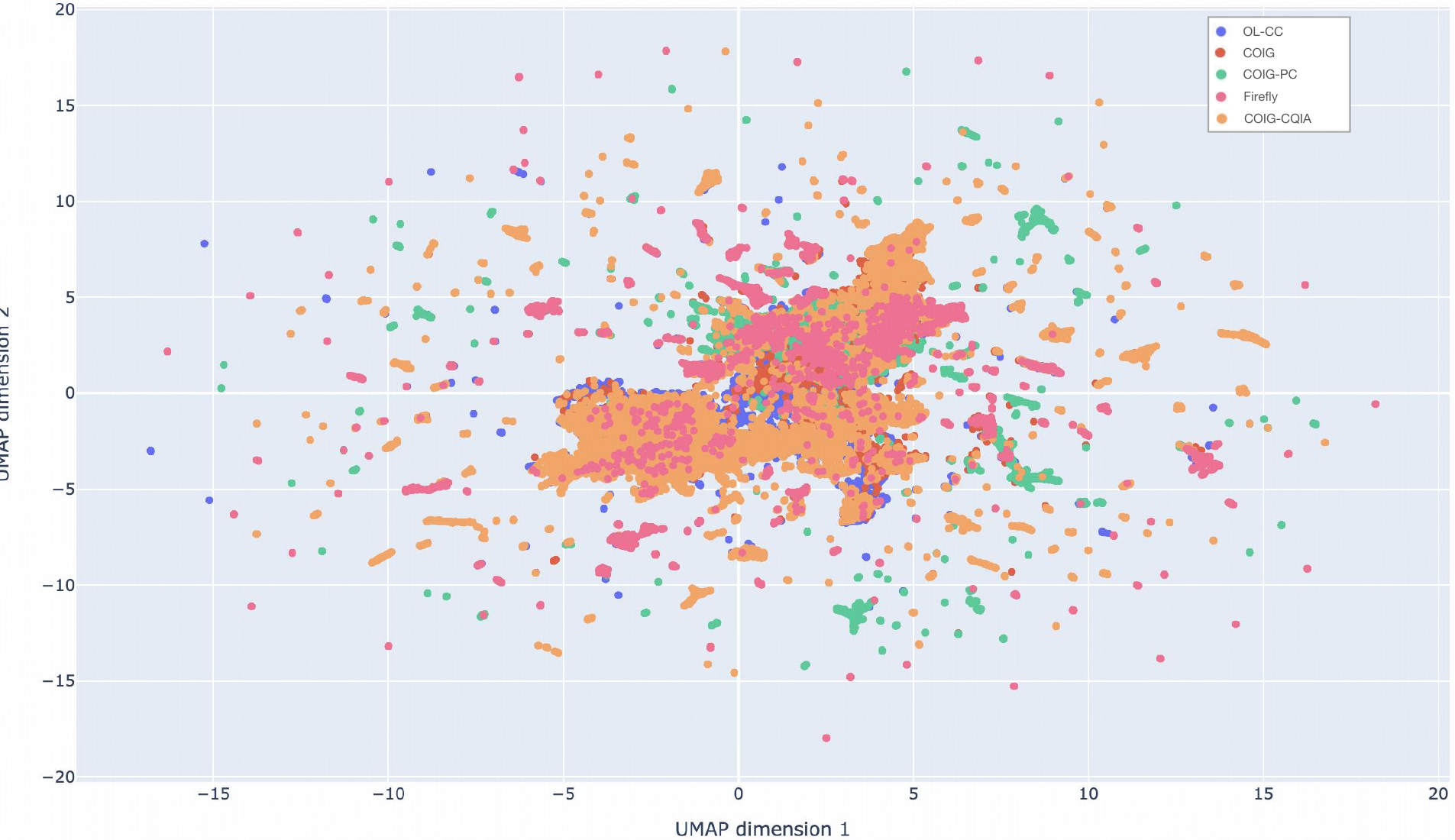}
\caption{U-Map Visualization of \data and other Chinese Datasets. \data exhibits the broadest distribution in the semantic space, encompassing the combined semantic distributions of all other datasets.}
\label{fig:u-map}
\end{figure*}

\section{Comparison between COIG-CQIA and other Chinese Datasets.}\label{appdix:comparison}

Table \ref{tab:comparison dataset} and Figure \ref{fig:u-map} show the comparison of \data and baseline Chinese Datasets.

% \newpage
\section{\data-Sub}\label{app:cqia-sub}

\begin{table}[htbp]
\centering
\resizebox{1\columnwidth}{!}{
\begin{tabular}{cccc}
\toprule
\textbf{Source} &  \textbf{Quantity} & \textbf{Source} & \textbf{Quantity}\\
\midrule
    Zhihu & 2733 & Douban & 300 \\
    Xiaohongshu & 50 & Segment Fault & 454\\
    Encyclopedia Article & 1350 & Encyclopedia of China & 200\\
    WikiHow & 300 & COIG PC & 3000 \\
    Middle school Exam & 200 & Graduate Entrance Examination & 475 \\
    Logi QA & 422 & CValue & 906 \\
    COIG-Human-Value & 101 & Chinese Traditional & 1110 \\
    Finance NLP Task & 500 & Ruozhiba & 240 \\
    Medical Article & 186 & Law & 400 \\
\midrule

    Total & & 12687 \\
\bottomrule 

\end{tabular}
}
\vspace{1em}
\caption{\label{tab:subset_source_statistics}
The data composition of CQIA-Subset.
% \byl{resize font}
}
\end{table}
To enrich the data diversity for CQIA, we have devoted to enlarging the coverage of data sources.
% In the full set of CQIA, we have devoted to expanding the coverage and scale of our  more data sources 
% collecting and annotating 
% have included as much data as we could collect and filter to facilitate the research or application requirements of the community. 
However, this lead to an imbalance distribution of data sources (e.g, Encyclopedia data accounts for a significant portion of the CQIA dataset).
% which could potentially reduce the diversity of the dataset. 
To alleviate this issue, we manually selected a high-quality subset, namely \data-Sub, which holds a more balanced data composition compared to the full set. Table \ref{tab:subset_source_statistics} shows the composition of CQIA-Subset.

\section{Experimental settings of evaluation}\label{app:detail eval}

For rapid evaluation, we selected an average of 200 samples from the BELLE-eval dataset based on task type to serve as our test set. Through our validation, we found that these 200 samples have a strong correlation with the full BELLE-eval dataset when used for model evaluation.
In human acessment, for each prompt, we generated one response per model\footnote{Responses were generated using nucleus sampling with $p$=0.85, $k$=50, and temperature=0.9.}, then asked annotators to compare the responses from our model and a baseline, allowing for a “tie” when neither response was better.

\begin{table}[h]
\centering
\resizebox{0.6\columnwidth}{!}{
\begin{tabular}{lc}
\toprule
\textbf{Model} & \textbf{SafetyBench} \\ 
\midrule
GPT-4-0613 & 89.2 \\
GPT-3.5-turbo-0613 & 80.4 \\
\midrule
Yi-6B & \\
+COIG PC & 81.2 \\
+Chinese Traditional & 76.6 \\
+Douban & 76.2 \\
+Exam & 77.6 \\
+Finance & 75.1 \\
+Logi QA & 79.1 \\
+Ruozhiba & 81.3 \\
+Segmentfault & 78.0 \\
+Wiki & 75.8 \\
+Wikihow & 76.4 \\
+Xhs & 76.0 \\
+Zhihu & 75.8 \\
+Human Value & 79.1 \\
\midrule
+\data & \textbf{81.7} \\
\bottomrule
\end{tabular}}
\caption{\label{tab:safetybench}SafetyBench scores of Yi-6B trained on various data sources.}\label{tab:safety}
\end{table}

% \section{Statistics}\label{appdix:stat}

\begin{table}[htbp]
\centering
\resizebox{\columnwidth}{!}{
\begin{tabular}{p{6cm}c}
\toprule
\textbf{Criteria} & \textbf{Accept Rate} \\
\midrule
\textit{Is the output correct and an acceptable answer?} & 98\% \\
\textit{Does the output meet the instructional requirements and provide a comprehensive and appropriate response to the question?} & 96\% \\
\textit{Is the answer complete and sufficiently detailed?} & 95\% \\
\textit{Is the answer harmless, avoiding misleading information or the spread of harmful content?} & 99\% \\
\bottomrule
\end{tabular}}
\caption{Accept Rate by Criteria.}\label{tab:acc_rate}
\end{table}

\section{Safety Evaluation}\label{app:safe}

Table \ref{tab:safetybench} shows that model trained on \data outperforms GPT-3.5-turbo-0613. 
Models trained on social media and forum data (e.g., Douban, Zhihu, and Xiaohongshu) achieved moderate safety scores, likely due to the diverse and open nature of social media content, which may cause potential harmfulness. 
Interestingly, models trained on Wiki-style data tended to score lower. We hypothesize that this may be due to the limited diversity of instruction within professional data sources, leading to poor performance on safety which is outside of specialized domains.

\end{CJK}
\end{document}